\documentclass{article}
\usepackage{spconf,amsmath,graphicx}

\usepackage{enumitem}
\setlist{nosep, leftmargin=14pt}

\usepackage{cite}
\usepackage{amsmath,amssymb,amsfonts}
\usepackage{hyperref}       
\usepackage{xcolor}

\DeclareFontFamily{U}{futm}{}
\DeclareFontShape{U}{futm}{m}{n}{
  <-> s * [.92] fourier-bb
  }{}
\DeclareSymbolFont{Ufutm}{U}{futm}{m}{n}
\DeclareSymbolFontAlphabet{\mathbb}{Ufutm}

\begin{document}

\newcommand{\ssmmodel}{Image2SSM}

\newcommand\ShapeRepresentation{Radial-Basis-Function Shape}
\newcommand\shaperepresentation{radial basis function shape}
\newcommand\srepresentation{RBF-shape}
\newcommand\samploss{narrow band loss}

\newcommand\Real{\mathbb{R}}
\newcommand\diff{\mathbf{d}}
\newcommand\aP{\mathbf{\mathcal{P}}}
\newcommand\aN{\mathbf{\mathcal{N}}}
\newcommand\aD{\mathbf{\mathcal{D}}}
\newcommand\bN{\mathbf{N}}

\newcommand{\SE}[1]{{\color{blue}#1}}
\newcommand{\HX}[1]{{\color{orange}#1}}
\newcommand{\TODO}[1]{{\color{red}#1}}
\newcommand{\EDIT}[1]{{\color{purple}#1}}

\def\x{{\mathbf x}}
\def\L{{\cal L}}

\newcommand{\var}{{\rm var}}
\newcommand{\Tr}{^{\rm T}}
\newcommand{\vtrans}[2]{{#1}^{(#2)}}
\newcommand{\kron}{\otimes}
\newcommand{\schur}[2]{({#1} | {#2})}
\newcommand{\schurdet}[2]{\left| ({#1} | {#2}) \right|}
\newcommand{\had}{\circ}
\newcommand{\diag}{{\rm diag}}
\newcommand{\invdiag}{\diag^{-1}}
\newcommand{\rank}{{\rm rank}}
\newcommand{\nullsp}{{\rm null}}
\newcommand{\tr}{{\rm tr}}
\newcommand{\vech}{{\rm vech}}
\renewcommand{\det}[1]{\left| #1 \right|}
\newcommand{\pdet}[1]{\left| #1 \right|_{+}}
\newcommand{\pinv}[1]{#1^{+}}
\newcommand{\erf}{{\rm erf}}
\newcommand{\hypergeom}[2]{{}_{#1}F_{#2}}

\renewcommand{\a}{{\bf a}}
\renewcommand{\b}{{\bf b}}
\renewcommand{\c}{{\bf c}}
\renewcommand{\d}{{\rm d}}  
\newcommand{\e}{{\bf e}}
\newcommand{\f}{{\bf f}}
\newcommand{\g}{{\bf g}}
\newcommand{\h}{{\bf h}}
\renewcommand{\k}{{\bf k}}
\newcommand{\m}{{\bf m}}
\newcommand{\mb}{{\bf m}}
\newcommand{\n}{{\bf n}}
\renewcommand{\o}{{\bf o}}
\newcommand{\p}{{\bf p}}
\newcommand{\q}{{\bf q}}
\renewcommand{\r}{{\bf r}}
\newcommand{\s}{{\bf s}}
\renewcommand{\t}{{\bf t}}
\renewcommand{\u}{{\bf u}}
\renewcommand{\v}{{\bf v}}
\newcommand{\w}{{\bf w}}
\newcommand{\y}{{\bf y}}
\newcommand{\z}{{\bf z}}
\newcommand{\A}{{\bf A}}
\newcommand{\B}{{\bf B}}
\newcommand{\C}{{\bf C}}
\newcommand{\D}{{\bf D}}
\newcommand{\E}{{\bf E}}
\newcommand{\F}{{\bf F}}
\newcommand{\G}{{\bf G}}
\renewcommand{\H}{{\bf H}}
\newcommand{\I}{{\bf I}}
\newcommand{\J}{{\bf J}}
\newcommand{\K}{{\bf K}}
\renewcommand{\L}{{\bf L}}
\newcommand{\M}{{\bf M}}
\newcommand{\N}{\mathcal{N}}  
\renewcommand{\O}{{\bf O}}
\renewcommand{\P}{{\bf P}}
\newcommand{\Q}{{\bf Q}}
\newcommand{\R}{{\bf R}}
\renewcommand{\S}{{\bf S}}
\newcommand{\T}{{\bf T}}
\newcommand{\U}{{\bf U}}
\newcommand{\V}{{\bf V}}
\newcommand{\W}{{\bf W}}
\newcommand{\X}{{\bf X}}
\newcommand{\Y}{{\bf Y}}
\newcommand{\Z}{{\bf Z}}

\newcommand{\bfLambda}{\boldsymbol{\Lambda}}

\newcommand{\bsigma}{\boldsymbol{\sigma}}
\newcommand{\balpha}{\boldsymbol{\alpha}}
\newcommand{\bpsi}{\boldsymbol{\psi}}
\newcommand{\bphi}{\boldsymbol{\phi}}
\newcommand{\boldeta}{\boldsymbol{\eta}}
\newcommand{\Beta}{\boldsymbol{\eta}}
\newcommand{\btau}{\boldsymbol{\tau}}
\newcommand{\bvarphi}{\boldsymbol{\varphi}}
\newcommand{\bzeta}{\boldsymbol{\zeta}}

\newcommand{\blambda}{\boldsymbol{\lambda}}
\newcommand{\bLambda}{\mathbf{\Lambda}}
\newcommand{\bOmega}{\mathbf{\Omega}}
\newcommand{\bomega}{\mathbf{\omega}}
\newcommand{\bPi}{\mathbf{\Pi}}

\newcommand{\btheta}{\boldsymbol{\theta}}
\newcommand{\bpi}{\boldsymbol{\pi}}
\newcommand{\bxi}{\boldsymbol{\xi}}
\newcommand{\bSigma}{\boldsymbol{\Sigma}}

\newcommand{\bgamma}{\boldsymbol{\gamma}}
\newcommand{\bGamma}{\mathbf{\Gamma}}

\newcommand{\bmu}{\boldsymbol{\mu}}
\newcommand{\1}{{\bf 1}}
\newcommand{\0}{{\bf 0}}

\newcommand{\bs}{\backslash}
\newcommand{\ben}{\begin{enumerate}}
\newcommand{\een}{\end{enumerate}}

 \newcommand{\notS}{{\backslash S}}
 \newcommand{\nots}{{\backslash s}}
 \newcommand{\noti}{{\backslash i}}
 \newcommand{\notj}{{\backslash j}}
 \newcommand{\nott}{\backslash t}
 \newcommand{\notone}{{\backslash 1}}
 \newcommand{\nottp}{\backslash t+1}

\newcommand{\notk}{{^{\backslash k}}}
\newcommand{\notij}{{^{\backslash i,j}}}
\newcommand{\notg}{{^{\backslash g}}}
\newcommand{\wnoti}{{_{\w}^{\backslash i}}}
\newcommand{\wnotg}{{_{\w}^{\backslash g}}}
\newcommand{\vnotij}{{_{\v}^{\backslash i,j}}}
\newcommand{\vnotg}{{_{\v}^{\backslash g}}}
\newcommand{\half}{\frac{1}{2}}
\newcommand{\msgb}{m_{t \leftarrow t+1}}
\newcommand{\msgf}{m_{t \rightarrow t+1}}
\newcommand{\msgfp}{m_{t-1 \rightarrow t}}

\newcommand{\proj}[1]{{\rm proj}\negmedspace\left[#1\right]}
\newcommand{\argmin}{\operatornamewithlimits{argmin}}
\newcommand{\argmax}{\operatornamewithlimits{argmax}}

\newcommand{\dif}{\mathrm{d}}
\newcommand{\abs}[1]{\lvert#1\rvert}
\newcommand{\norm}[1]{\lVert#1\rVert}

\newcommand{\ie}{{{i.e.,}}\xspace}
\newcommand{\eg}{{{\em e.g.,}}\xspace}
\newcommand{\EE}{\mathbb{E}}
\newcommand{\VV}{\mathbb{V}}
\newcommand{\sbr}[1]{\left[#1\right]}
\newcommand{\rbr}[1]{\left(#1\right)}
\newcommand{\cmt}[1]{}

\title{Optimization-Driven Statistical Models of Anatomies using Radial Basis Function Shape Representation}


\name{Hong Xu and Shireen Y. Elhabian\sthanks{The National Institutes of Health supported this work under grant numbers NIBIB-U24EB029011 and NIAMS-R01AR076120. The content is solely the responsibility of the authors and does not necessarily represent the official views of the National Institutes of Health.}}
\address{Scientific Computing and Imaging Institute, Kahlert School of Computing, \\ University of Utah, Salt Lake City, UT, USA \\ \{hxu,shireen\}@sci.utah.edu }


\maketitle

\begin{abstract}
Particle-based shape modeling (PSM) is a popular approach to automatically quantify shape variability in populations of anatomies. The PSM family of methods employs optimization to automatically populate a dense set of corresponding particles (as pseudo landmarks) on 3D surfaces to allow subsequent shape analysis. A recent deep learning approach leverages implicit radial basis function representations of shapes to better adapt to the underlying complex geometry of anatomies. Here, we propose an adaptation of this method using a traditional optimization approach that allows more precise control over the desired characteristics of models by leveraging both an eigenshape and a correspondence loss. Furthermore, the proposed approach avoids using a black-box model and allows more freedom for particles to navigate the underlying surfaces, yielding more informative statistical models. We demonstrate the efficacy of the proposed approach to state-of-the-art methods on two real datasets and justify our choice of losses empirically. 
\end{abstract}

\begin{keywords}
Statistical Shape Modeling, Optimization, Radial Basis Function Interpolation, Polyharmonic Splines.
\end{keywords}

\section{Introduction}

Statistical Shape Modeling (SSM) is a family of techniques used to quantify shape variability given a population of anatomies. SSM has advanced the understanding of different diseases and disorders \cite{atkins2017quantitative, bhalodia2020quantifying, LenzTalocruralJoint, VANBUURENHipOsteoarthritis, merle2019high,bruse2016statistical}.
%
%
SSM relies on representing shape variation in a compact way to perform subsequent statistical analyses. The two most popular shape representations used for building SSMs are \textit{deformation fields} and \textit{landmarks}. Deformation fields encode \textit{implicit} transformations between population samples and a pre-defined (or learned) atlas. Landmarks are \textit{explicit} points spread on shape surfaces that correspond across the population and serve as proxies to the surface for subsequent statistical analysis\cite{sarkalkan2014statistical}. 
Shape analysis used to be performed by manually placing landmarks and performing statistics on them, but this was labor-intensive, not reproducible, and required extensive domain expertise (e.g., radiologists).
Modern computational methods allow the automatic computation of deformations or placement of landmarks, shifting the SSM field to data-driven characterization of population-level variabilities that is objective, reproducible, and scalable. These methods include minimum description length -- MDL \cite{davies2002minimum}, particle-based shape modeling -- PSM \cite{cates2017shapeworks} for landmarks, and frameworks based on Large Deformation Diffeomorphic Metric Mapping for deformation fields (e.g., \cite{durrleman2014morphometry}). 
%

Landmark-based representations, also known as point distribution models (PDMs), have been favored for analyzing surface variability due to their simplicity, computational efficiency, and interpretability for statistical analyses \cite{sarkalkan2014statistical}. 
However, such scalability suffers when considering that some applications require thousands of densely placed landmarks to capture shape structure faithfully when representing intricate shape surfaces with localized or convoluted shape features that may live between landmarks. The state-of-the-art PDM method, PSM \cite{cates2017shapeworks}, produces dense particle placements. However, it provides a uniform surface sampling, and thus fail to adapt to the underlying geometrical intricacies at different scales.


Image2SSM \cite{Image2SSM} first proposed the use of radial basis function shape representations (RBF-shapes) for the SSM task. 
RBF-shapes are continuous surface representations instead of simply discrete surface proxies. They allow particles to self-adapt to poorly represented surface areas while building the statistical model. 
Despite its benefits, the black-box nature of such a deep network optimization undermines the user's ability to understand and control the process of building SSMs, not to mention the computational cost of parameterizing the PDM construction with a deep network. Furthermore, parameterizing landmarks using a network limits particles manipulation 
at both training and inference times, leading to poor surface adherence and configuration congruency. 

To address these limitations,
this paper proposes a non-deep optimization-based solution that leverages RBF-shapes to build SSMs.
%
%
It has the advantage of being self-adaptive to the underlying geometry and can further reduce the number of required particles for intricate surfaces. Additionally, not using a black box model deep learning approach gives our method much more control over the particle spread and placement since they are not parametrized with a deep neural network. We also introduce the simultaneous use of correspondence and eigenshape losses to improve configuration consistency while emphasizing dominant modes of variation. 
Additionally, enhancing traditional optimization techniques contributes to the advancement of the deep learning variants for PDM construction (e.g., \cite{bhalodia2018deepssm, uncertaindeepssm, adams2022images,adams2023fully}) by providing stronger shape priors to train the networks. 
We show that our method performs similarly or better than PSM  \cite{cates2017shapeworks} and Image2SSM \cite{Image2SSM} 
on 
real datasets in 
correspondence metrics and two-way surface-to-surface distance. We also show how the correspondence and eigenshape losses function to build better PDMs. 
The proposed method will be integrated into the open-source package for building PDMs, ShapeWorks \cite{cates2017shapeworks}.
\section{Methods}


A statistical shape model should balance an accurate geometrical representation of the individual surfaces in the given cohort while preserving landmark configuration congruency across the surfaces, referred to as \textit{correspondence}, encoding shape variations in a \textit{compact} set of parameters. 
%
The proposed method optimizes four losses using stochastic gradient descent to achieve such a balance by explicitly emphasizing and controlling the different characteristics of an SSM. 
The objective is to optimize for a set of landmarks (or control points in RBF literature). 
These control points are fit to a cohort of shapes in the form of segmentations $\mathcal{S} = \{\S_i\}_{i=1}^I$. We optimize for a set of $J$ control points $\aP = \{\P_i\}_{i=1}^I$ for each input shape, where the $i-$th shape PDM 
is denoted by $\P_i = [\p_{i,1}, \p_{i,2}, \cdots, \p_{i,J}]$ and $\p_{i,j} \in \mathbb{R}^3$. The RBF-shape can then reconstruct a full surface using these points and their surface normals by interpolating between them. This provides online information to the model regarding how well each part of the surface is represented, leading to adaptive updates.

This section will first explain how the RBF-shape can estimate reconstruction error at each region and how each of the four losses is involved in the optimization.



\subsection{Radial Basis Function Surface Representation}
\begin{figure}[!t]
\centerline{\includegraphics[width=7cm]{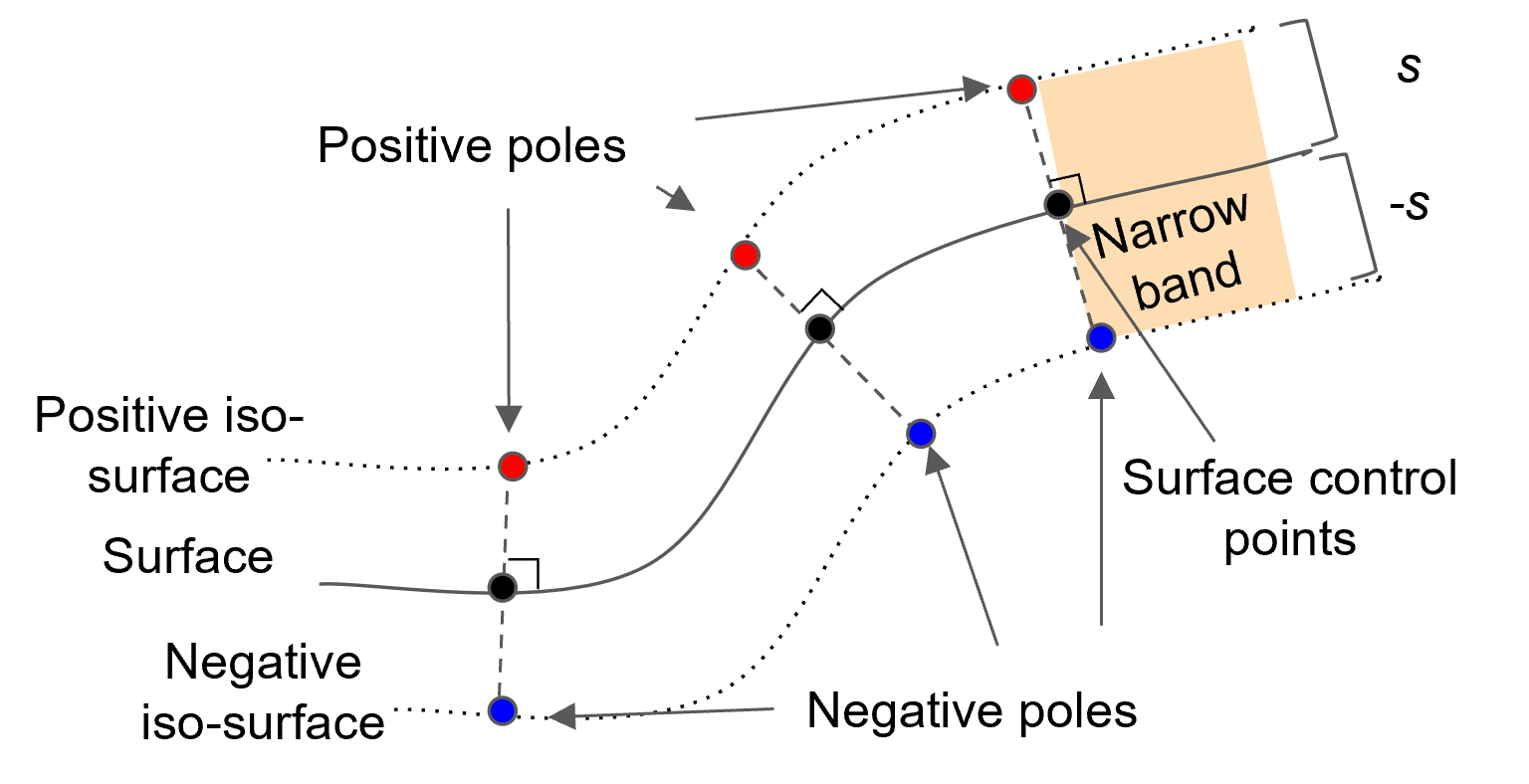}}
\vspace{-0.15in}
\caption{A set of control points and their respective dipoles can define an implicit signed distance function within a narrow band that can be efficiently queried for surface representation accuracy at any narrow band point. }
\label{dipole}
\end{figure}

\textit{\srepresentation}\ is an implicit surface representation based on radial basis functions, which can represent intricate shapes by leveraging both surface control points and normals to inform shape reconstructions \cite{Turk1999VariationalIS, rbf_surface1}. It uses a set of control points at the zero-level set and a pair of off-surface points (aka \textit{dipoles}) with a signed distance $s$ and $-s$ along the surface normal of each control point. 
%
%
This is illustrated in Fig. \ref{dipole}. 

We refer to the set of control points and their dipoles as $\widetilde{\P_i}$ for shape $i$, where $\widetilde{\P_i} = [\P_i, \P_i^+, \P_i^-]$ with $\p_{i,j}^{\pm} = \p_{i,j} \pm s\n_{i,j}$.
%
%
The shape's implicit function, which maps a point $\x \in \mathbb{R}^3$ to its signed distance to the surface can then be written as: 


\vspace{-0.18in}
\begin{equation} \label{eq:rbf}
\begin{split}
    f_{\widetilde{\P_i}, \w_i}(\x) = \sum_{j \in \widetilde{\P_i}} w_{i,j} \phi(\x, \widetilde{\p}_{i,j}) + \mathbf{c}^T_i \x + c^0_{i} 
\end{split}
\vspace{-0.2in}
\end{equation}

\noindent where $\phi$ can be any RBF basis function (e.g., the thin plate spline $\phi(\x,\y) =(|\x-\y|_2)^2\log(|\x-\y|_2)$, the biharmonic  $\phi(\x,\y) =|\x-\y|_2$ or the triharmonic $\phi(\x,\y) =(|\x-\y|_2)^3$) and $\mathbf{c}_i \in \mathbb{R}^3$ and $c^0_{i} \in \mathbb{R}$ encode the linear component of the surface. We obtain $\w_i = [w_{i,1}, w_{i,2}, ..., w_{i,3J}, c_i^0, c_i^1, c_i^2,c_i^3] \in \mathbb{R}^{3J + 4}$ by solving a system of equations formed by Eq \ref{eq:rbf} over $\x \in \widetilde{\P_i}$ and regularization factors to form a fully determined system,
fully described in \cite{Turk1999VariationalIS, rbf_surface1}. 
The function $f$ provides an implicit shape representation that is resolution agnostic and can be queried to construct segmentations and surface meshes at arbitrary resolution. 
This representation excels at adapting to the underlying surface geometry, leading to better shape representation using fewer control points.




\subsection{SSM Optimization} 
Control points for each surface are initialized using the same set of points sampled uniformly at random.
During the optimization, the four losses 
ensure that the control points (1) live on the surface, (2) maintain the same configuration across all shapes, (3) push the population-level variability to dominant modes of variation, and (4) are spread according to the reconstruction error of the RBF-shape representation. 
Together, these losses promote a balance between the geometric representation of the individual surfaces and maintaining consistent placement of control points across the population while resulting in a compact statistical model.

\vspace{0.1in}
\noindent \textbf{Surface loss:} The surface loss aims to force control points to lie on the surface. For each shape $i \in I$, we define this loss as


\vspace{-0.2in}
\begin{equation}
\begin{split}
  L^{surf}_{\D_i} (\P_i) &= \sum_{j = 1}^J  |\D_i(\p_{i,j})|,
\end{split}
\vspace{-0.2in}
\end{equation}

\noindent where $\D_i(\p_{i,j})$ is the true distance to the closest surface point to $\p_{i,j}$, obtained by querying the signed distance transform.



\vspace{0.1in}
\noindent \textbf{Correspondence loss:} The notion of control points correspondence across the shape population can be quantified by the Frobenius norm of the covariance of the control points. In simple terms, this loss attempts to collapse each control point to its mean across all shapes. It is written as 
%
\vspace{-0.12in}
\begin{equation}
\begin{split}
  L^{corres}_{\boldsymbol{\mu}} (\P_1, ..., \P_K) & 
  = \left\|\sum_{k=1}^K \left(\P_k - \boldsymbol{\mu}\right) \left(\P_k - \boldsymbol{\mu}\right)^T \right\|_F
\end{split}
\vspace{-0.25in}
\end{equation}
\noindent where 
$\boldsymbol{\mu} = \sum_{i=1}^I \P_i$ is the mean control points, and $\|\cdot\|_F$ the Frobenius norm.

\vspace{0.1in}
\noindent \textbf{Eigenshape loss:} This loss pushes control points to align with the most dominant modes of variability. In \cite{cates2017shapeworks}, this loss is the one used as the correspondence loss (derived from \cite{EigenshapeKOTCHEFF1998303}). This objective does not improve correspondence of the control points, but rather improves compactness by aligning control points to the principal components. 
This observation aligns with the source claims about this loss in \cite{EigenshapeKOTCHEFF1998303}, which described this objective as being equivalent to minimizing the Mahalanobis distance of the system. However, due to the eigenvalue scaling inherent in the Mahalanobis distance definition, missed correspondences get captured by the principal components and are not corrected by this loss, making it a poor loss to enforce correspondence. This shortcoming is shown empirically in our results section, hence our choice of using the Frobenius norm loss to enforce correspondence. Nonetheless, including this loss is imperative to improving the compactness of the resulting shape model, which is a desirable property in the analysis that does not severely interfere with good sampling and correspondence.

Given a minibatch of size $K$, the eigenshape loss is formulated using the differential entropy $H$ of the samples in the minibatch, assuming a Gaussian distribution.

\vspace{-0.2in}
\begin{equation}
\begin{split}
  L^{eigen}_{\boldsymbol{\mu}} (\P_1, ..., \P_K) & = H(\P)  \\
   = \frac{1}{2} \log \left| \frac{1}{3JK}\sum_{k=1}^K \right.& \left. \left(\P_k - \boldsymbol{\mu}\right) \left(\P_k - \boldsymbol{\mu}\right)^T \right|
\end{split}
\vspace{-0.2in}
\end{equation}
\noindent where $\P$ here indicates the random variable of the shape space and $|\cdot|$ is the matrix determinant.
Both the correspondence and eigenshape loss are triggered starting from the second epoch as soon as a mean particle set $\boldsymbol{\mu}$ is available.




\vspace{0.1in}
\noindent \textbf{Sampling loss:} This loss encourages the control points to adapt to the underlying geometry. We query the distance at $R$ randomly sampled points $\B_i = [\b_{i,1}, ..., \b_{i,R}]$ that lie within a \textit{narrow band} at $\pm s$ distance from the surface. The control points are then guided to poorly represented areas by minimizing their distance to the \textit{surrounding} narrow band points, scaled by the queried approximate violation intensity.

%
%

To define surrounding narrow band points, let $\K^i \in \mathbb{R}^{R \times M}$ be the matrix containing the pairwise distances between each narrow band point $\b_{i,r}$ and each control point $\p_{i,j}$, such that, for the $i-$th shape, the $r,j-$th element is $k_{r,j}^i = \|\b_{i,r} - \p_{i,j}\|_2$. We use a soft minimum function to allow each narrow band point to influence all nearby control points, written as $\operatorname{softmin}(\K^i)$, where the $r,j$th element of $\operatorname{softmin}(\K^i)$ is computed as $\exp{(-k_{r,j}^i)} / \sum_{j'=1}^J \exp (k_{r,j'}^i)$. This soft minimum is normalized over $\P_i$. Finally, the RBF approximation squared error at the narrow band points is captured by $\e_i \in \mathbb{R}_+^{R}$, where $e_{i,r} = [f_{\widetilde{\P_i}, \w_i}(\b_{i,r}) - \D_i(\b_{i,r})]^2$. Let $\E_i = \e_i \mathbf{1}_M^T$, where $\mathbf{1}_M$ is a ones-vector of size $M$. The sampling loss is then written as 

\vspace{-0.2in} 
\begin{equation}
\begin{split}
  L^{sampl}_{\B_i, \D_i, \w_i} (\P_i, \bN_i) & = \\
  \operatorname{mean} & \left(\operatorname{softmin}(\K_i) \otimes \K_i \otimes  (c\E_i + 1) \right)
\end{split}
\vspace{-0.2in} 
\end{equation}
\noindent where $\otimes$ indicates the Hadamard (elementwise) multiplication of matrices, $c$ is a weighting variable, and $\operatorname{mean}$ computes the average over all the matrix elements. Adding 1 to $\E_i$ gives each narrow band point at least a unit pull.






\vspace{0.1in}
\noindent \textbf{Total loss:} We optimize minibatches of size $K$ at a time with a total loss written as follows:


\vspace{-0.3in} 
\begin{equation}
\begin{split}
  L_{\mathcal{I}, \aD, \partial \aD}(\aP_K, \aN_K) &= \sum_{i=1}^K \biggl( \alpha L^{surf}_{\D_i} (\P_i) \\
  & + \beta L^{sampl}_{\B_i, \D_i, \w_i} (\P_i, \bN_i) \biggl) \\
  & +  \gamma L^{eigen}_{\boldsymbol{\mu}} (\P_1, ..., \P_K)\\
  & + \zeta L^{corres}_{\boldsymbol{\mu}} (\P_1, ..., \P_K)
\end{split}
\vspace{-0.2in} 
\end{equation}


\noindent where $\alpha,\beta, \gamma$, and $\zeta \in \mathbb{R}_+$  are hyperparameters of each losses and $\aP_K, \aN_K$ are the control points and their surface normals of the samples in the minibatch. 

\vspace{-0.15in}
\section{Results}

Our results demonstrate that the proposed method outperforms Image2SSM \cite{Image2SSM} in both two-way surface-to-surface distance and correspondence metrics while improving slightly over PSM \cite{cates2017shapeworks}. Additionally, we showcase the effect of the correspondence loss and the eigenshape loss on a single major axis variation ellipsoid toy example to justify our choice of loss functions.
\vspace{-0.15in}
\subsection{Datasets and Implementation Details}

We perform tests on two real datasets. First, a dataset of signed distance transforms from 40 proximal femur CT scans devoid of pathologies. 
The second dataset consists of 814 left atrium signed distance transforms, a challenging dataset due to size and high shape variability. 

We used the PyTorch 1.12.1 Autograd functionality to automatically backpropagate our losses using the SGD optimizer. As a preprocessing step, a set of uniformly sampled control points is used to initialize the optimization on a single shape.
These control points then serve as the initialization for all shapes during the optimization of all control points. This preprocessing is not strictly necessary as the main optimization can be initialized with the same random set of points, but it reduces the total number of iterations needed by leveraging this reasonable prior. This is in contrast to the PSM initialization, which utilizes a multi-scale particle splitting routine to initialize the shape model.

At each epoch, 10,000 narrow band points are sampled from the narrow band of each surface to accelerate computation. We used the biharmonic kernel for the basis function, but the results are observed to be agnostic to this choice. The hyperparameters we used are $learning\_rate = 1$, $\alpha = 0.1,\beta=1.0, \gamma=0.01$, $\zeta=0.01$, and $c=10$ for femurs; changing $\zeta=0.1$, $\gamma=0.001$, $\zeta=0.001$ for left atria; and $\alpha = 0.01$, and $c=1$ for ellipsoids.
\vspace{-0.15in}
\subsection{Shape Model Quality}

\begin{figure}[!t]
\centerline{\includegraphics[width=7cm]{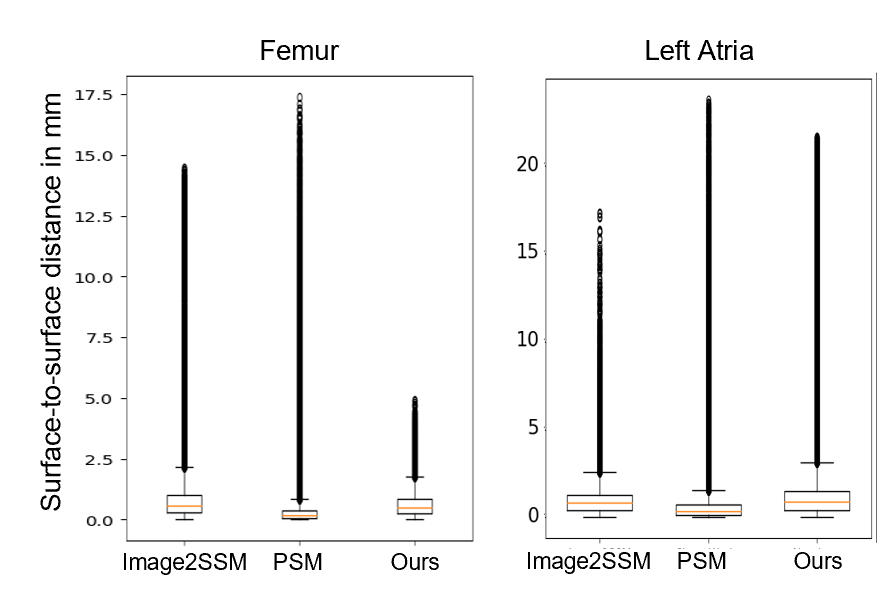}}
\vspace{-0.15in}
\caption{The two-way surface-to-surface distance between Image2SSM, PSM, and our approach.}
\label{boxplots}
\end{figure}

We evaluate our approach against the state-of-the-art Image2SSM \cite{Image2SSM} and PSM \cite{cates2017shapeworks} in building a model with 128 particles.
Fig. \ref{boxplots} shows that our approach matches or outperforms Image2SSM and PSM in the maximum two-way surface-to-surface distance of the reconstruction while yielding a slightly poorer mean distance than PSM. It resolves the worst surface misrepresentations but fails to capture the general surface as closely without an explicit snap-to-surface routine like PSM uses.

\begin{figure}[!t]
\centerline{\includegraphics[width=7.5cm]{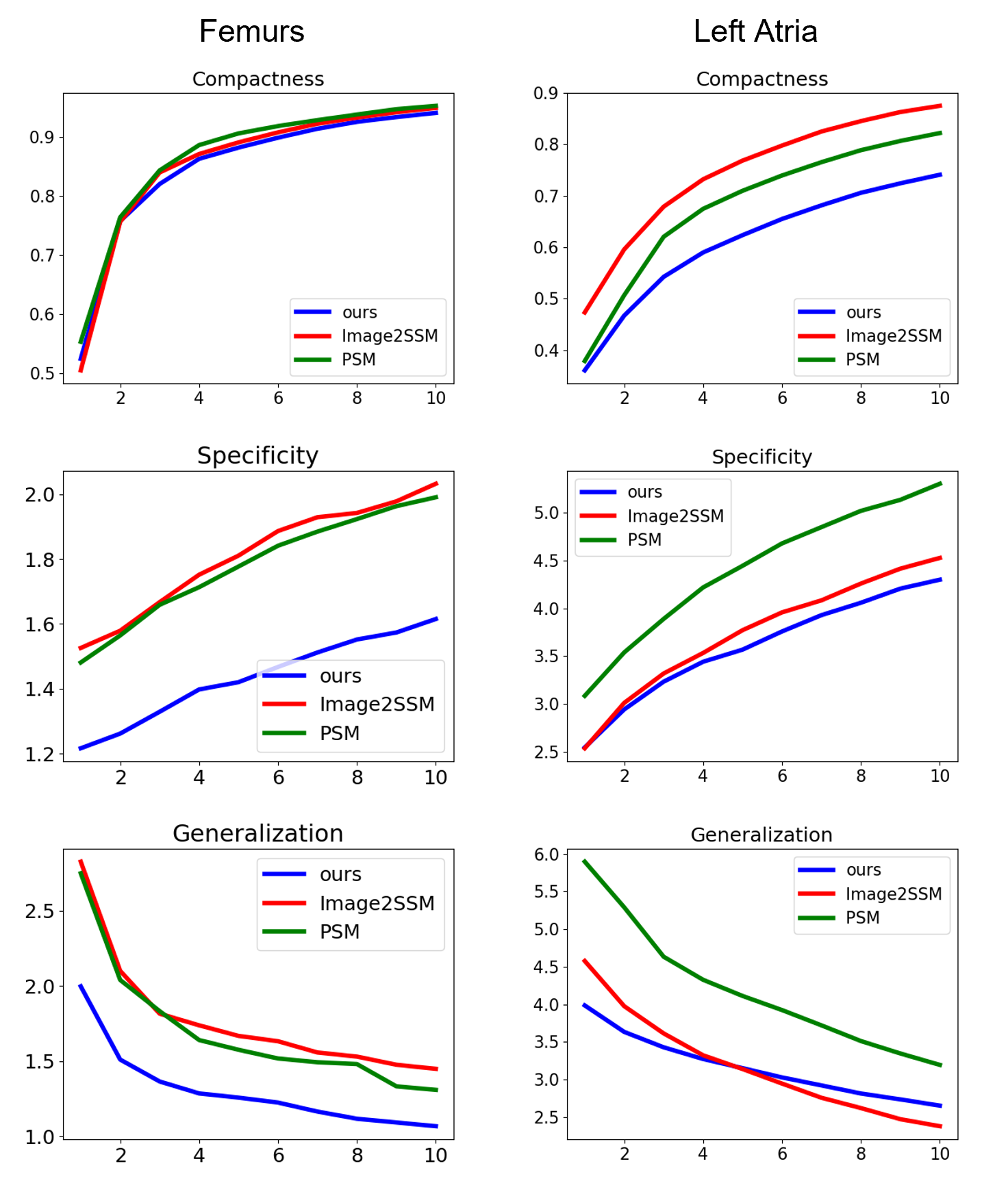}}
\vspace{-0.15in}
\caption{Shows the compactness (higher is better), specificity (lower is better), and generalization (lower is better) for ten modes on femurs and left atria. 
}
\label{metrics}
\end{figure}

Fig. \ref{metrics} shows conventional metrics of compactness (percentage of variance captured), specificity (ability to generate realistic shapes), and generalization (ability to represent unseen shape instances) \cite{Davies2004LearningSO} for femurs and left atria compared to Image2SSM and PSM for the first ten modes of variation, where it yields comparable compactness but better specificity and generalization.

\vspace{-0.15in}
\subsection{Correspondence and Eigenshape Losses}

\begin{figure}[!t]
\centerline{\includegraphics[width=8cm]{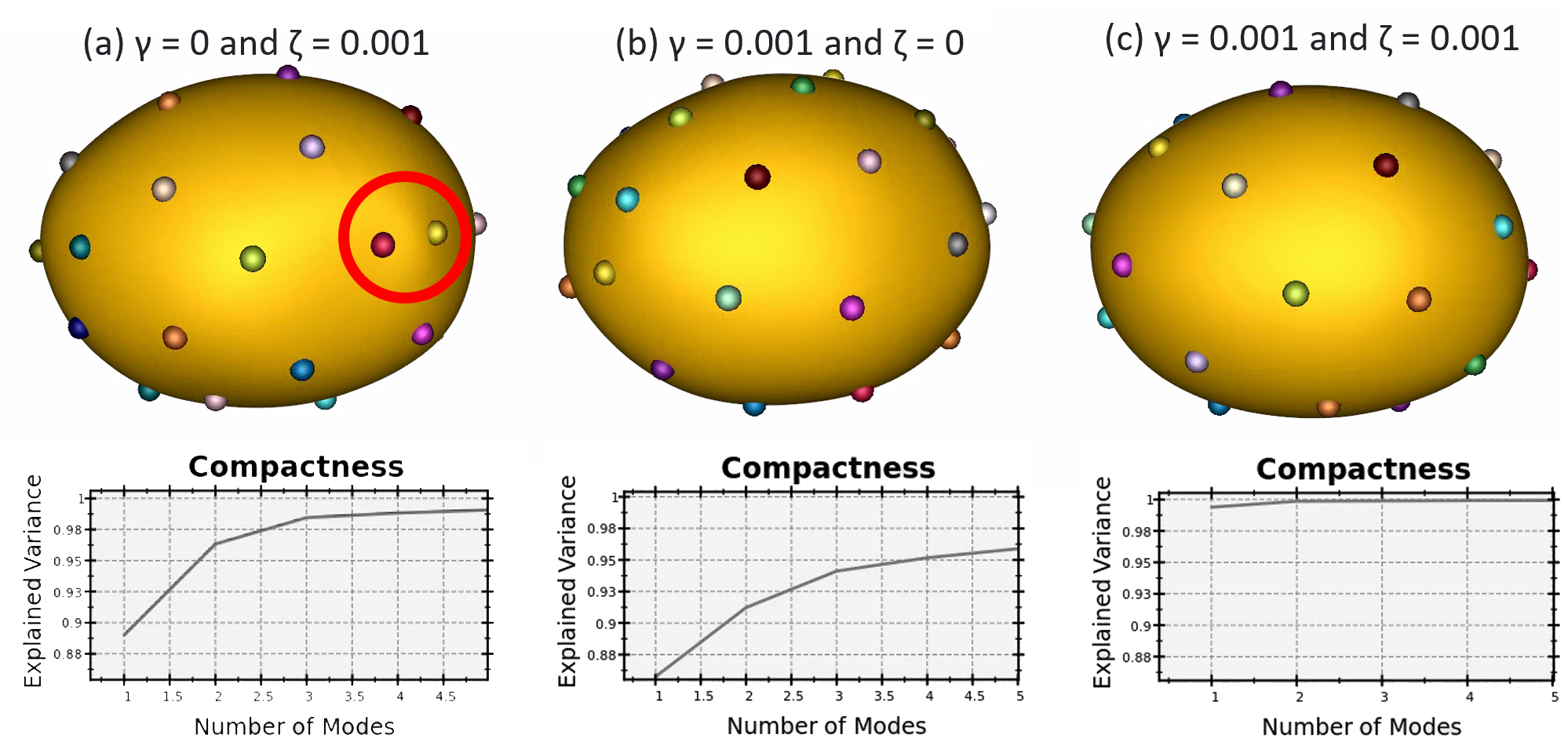}}
\vspace{-0.15in}
\caption{By changing $\gamma$ and $\zeta$, we show the results of running our optimization with (a) only correspondence loss, (b) only eigenshape loss, and (c) both. The red circle highlights two control points that are swapped for 6 out of 20 shapes.}
\label{eigen_corr}
\end{figure}

For simplicity, we show the effects of the correspondence and eigenshape loss on a population of ellipsoids where only the major axis $x$ varies. Fig. \ref{eigen_corr}(a) shows how a missed correspondence is not only not fixed by the correspondence loss, but is captured in its main mode of variation. Fig. \ref{eigen_corr}(b) shows how even with perfect correspondence, courtesy of the correspondence loss, the percentage variance of the main mode and compactness are worse than \ref{eigen_corr}(a) due to the lack of eigenshape loss. Fig. \ref{eigen_corr}(c) shows how we can achieve a compact model with good correspondences using both losses. All in all, this shows not only that the eigenshape loss is inadequate to resolve correspondences, but it can also generate spurious variability in the main modes of variation.

\vspace{-0.15in}
\section{Conclusion}
We demonstrate an approach for improving PDM construction using a traditional optimization technique. Our proposed losses offer new insights regarding the development of correspondence models that better capture variability while being faithful to the surface representations. Future work involves fully exploring the effect of existing losses for correspondence and how they interact against the eigenshape loss, as well as furthering RBF-shape's ability to capture surface error to enable surface-informed non-uniform sampling.

\section{Compliance with Ethical Standards}

This study was performed in line with the principles of the Declaration of Helsinki. Approval was granted by the Institutional Review Board at the University of Utah IRB\#00056086 for the femur dataset and IRB\#00072747 for the left atrium dataset.


\bibliographystyle{IEEEbib}
{\footnotesize \bibliography{references}}

\end{document}